\documentclass[letterpaper, 10 pt, conference]{ieeeconf}

\IEEEoverridecommandlockouts                              

\usepackage{epsfig} 
\usepackage{mathptmx} 
\usepackage{times} 
\usepackage{amsmath} 
\usepackage{amssymb}  

\usepackage{multirow}
\usepackage{graphicx} 
\usepackage{newfloat}
\usepackage{listings}
\usepackage{float}
\usepackage{cite}
\usepackage{booktabs}

\title{\LARGE \bf
CVFNet: Real-time 3D Object Detection by Learning Cross View Features
}

\author{    
        Jiaqi Gu\textsuperscript{\rm 1}, 
        Zhiyu Xiang\textsuperscript{\rm 2}, 
        Pan Zhao\textsuperscript{\rm 1}, 
        Tingming Bai\textsuperscript{\rm 1}, 
        Lingxuan Wang\textsuperscript{\rm 1}, 
        Xijun Zhao\textsuperscript{\rm 3}, 
        Zhiyuan Zhang\textsuperscript{\rm 2,4}
\thanks{$^{1}$Jiaqi Gu, Pan Zhao, Tingming Bai and Lingxuan Wang are with College of Information Science \& Electronic Engineering, Zhejiang University, Hangzhou, China, {\tt\footnotesize\{vadin, pan\_zhao, incredibai, 21960188\}@zju.edu.cn}}%
\thanks{$^{2}$Zhiyu Xiang, corresponding author, is with Zhejiang Provincial Key Laboratory of Information Processing, Communication and Networking, Zhejiang University, Hangzhou, China.{\tt\footnotesize xiangzy@zju.edu.cn}}%
\thanks{$^{3}$Xijun Zhao, is with China North Vehicle Research Institute, Beijing, china.{\tt\footnotesize heejunzhao@163.com}}%
\thanks{$^{4}$Zhiyuan Zhang, is with Ningbo Research Institute, Zhejiang University, Ningbo, China.{\tt\footnotesize cszyzhang@gmail.com}}%
}

\begin{document}

\maketitle
\thispagestyle{empty}
\pagestyle{empty}

\begin{abstract}
In recent years 3D object detection from LiDAR point clouds has made great progress thanks to the development of deep learning technologies. Although voxel or point based methods are popular in 3D object detection, they usually involve time-consuming operations such as 3D convolutions on voxels or ball query among points, making the resulting network inappropriate for time critical applications. On the other hand, 2D view-based methods feature high computing efficiency while usually obtaining inferior performance than the voxel or point based methods. In this work, we present a real-time view-based single stage 3D object detector, namely CVFNet to fulfill this task. To strengthen the cross-view feature learning under the condition of demanding efficiency, our framework extracts the features of different views and fuses them in an efficient progressive way. We first propose a novel Point-Range feature fusion module that deeply integrates point and range view features in multiple stages. Then, a special Slice Pillar is designed to well maintain the 3D geometry when transforming the obtained deep point-view features into bird's eye view. To better balance the ratio of samples, a sparse pillar detection head is presented to focus the detection on the nonempty grids. We conduct experiments on the popular KITTI and NuScenes benchmark, and state-of-the-art performances are achieved in terms of both accuracy and speed.
\end{abstract}

\section{Introduction}\label{sec:introduction}
3D object detection, which outputs a series of 3D bounding boxes specifying the size, 3D pose and class of the objects in the environment, is an essential perception task for autonomous vehicle. In recent years, LiDAR has become one of the most widely used sensors for autonomous vehicles because of its accurate distance measurement and the robustness to illumination changes. 
It therefore becomes imperative to design effective 3D object detectors to boost the autonomous navigation. 
However, efficiently detecting objects in 3D LiDAR point clouds still remains challenging due to the irregular format and the sparsity of the point cloud.

\begin{figure}[H]
   \setlength{\belowcaptionskip}{-0.6cm}
   \centering
   \includegraphics[width=8cm]{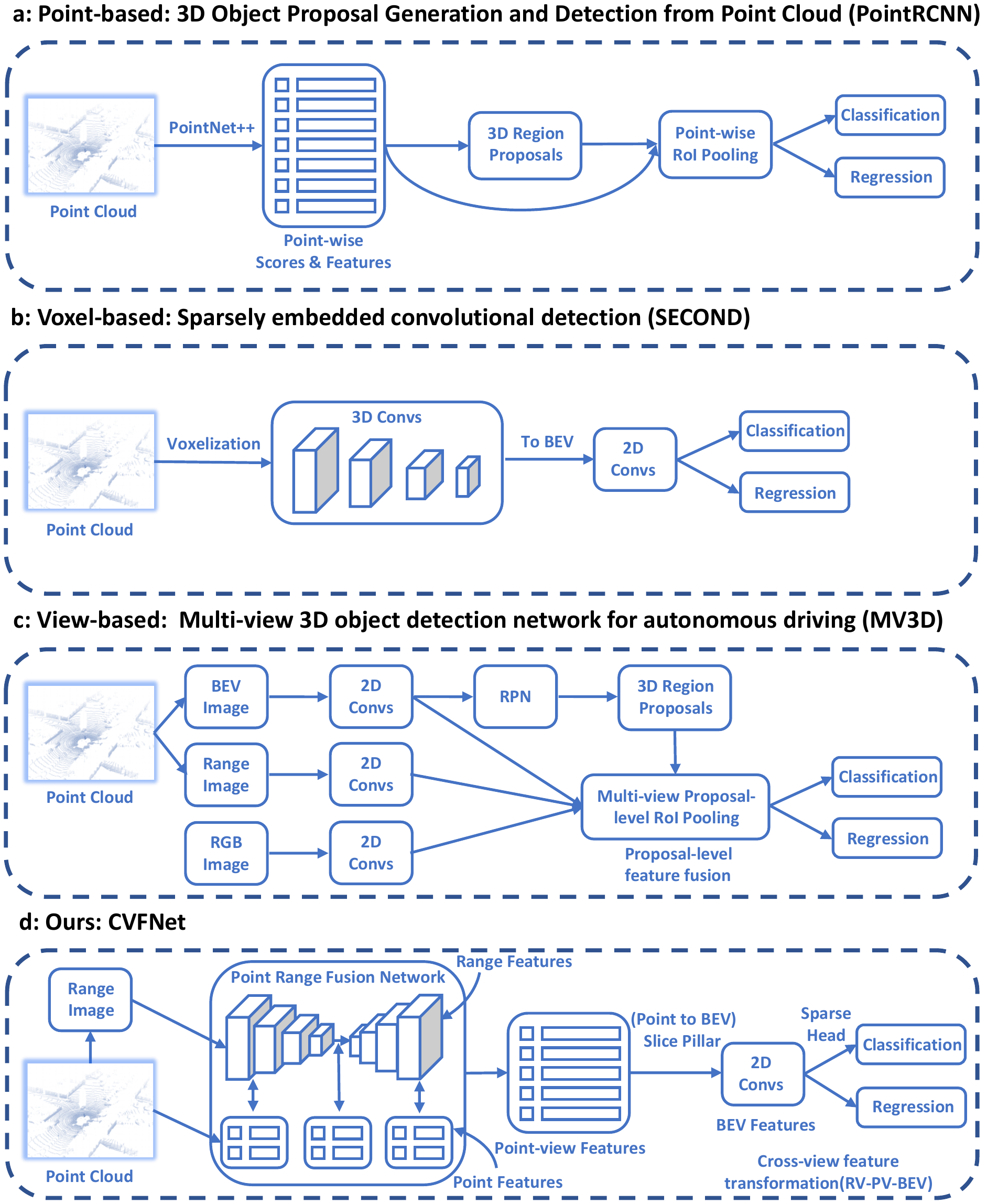}
   \caption{Comparison with state-of-the-art 3D object detection methods with different representations. Instead of using time-consuming PointNet++ or 3D Convolutions in (a), (b) and proposal-level feature aggregation in (c), our method rely on 2D convolutions and fuse the cross-view features in an effectively progressive way.}
   \label{fig:comparison}
\end{figure}

As Convolutional Neural Networks (CNN) have shown great advances on 2D object detection, some works try to adapt the success of CNN to 3D object detection by dividing the point cloud into voxels followed by 3D convolutions to learn the voxel-wise features~\cite{wang2015voting, song2016deep, Zhou2018VoxelNetEL, Yan2018SECONDSE, Chen2019FastPR}. Since LiDAR point clouds are sparse by nature, most computation is wasted on the redundant empty voxels, which makes these methods inefficient~\cite{wang2015voting, li2016vehicle, Zhou2018VoxelNetEL}. Although sparse convolutions~\cite{graham2017submanifold, graham20183d} can be employed for acceleration~\cite{Yan2018SECONDSE}, the efficiency and deployment are still inferior to the 2D counterpart~\cite{he2017mask, redmon2017yolo9000}.

Inspired by the PointNet~\cite{Qi_2017_CVPR} and its variant PointNet++~\cite{Qi2017PointNetDH} which pioneer the point-wise feature extraction, point-based 3D detection methods~\cite{Qi2018FrustumPF, Shi2019PointRCNN3O, Wang2019FrustumCS, yang2019std} are proposed to directly estimate the 3D bounding boxes from raw point cloud. While it is free of quantization loss compared to the voxel-based methods, the ball query operations for aggregating local features from neighboring points involved in PointNet++ are time-consuming. 
 
Another line are view-based works which project the whole point cloud to some 2D views (\textit{e}.\textit{g}. bird’s eye view (\textbf{BEV})~\cite{Chen2017Multiview3O, Yang2018PIXORR3} or range view~\cite{Liang2020RangeRCNNTF, liang2021rangeioudet}) and rely on 2D convolutions to extract features. They usually extract features from a single view image or concatenate different view features together to accomplish the detection. These methods attract increasing attention in recent years due to its high computing efficiency. However, view-based approaches usually suffer from the information loss during the projection from 3D points to 2D views, which is hard to be compensated during 3D detection.  When multiple views are employed, directly concatenating features in proposal level as in MV3D\cite{Chen2017Multiview3O} can weaken the consistency among features. These shortcomings lead to inferior performance of the view-based methods than the voxel or point based counterparts.

In this work, we propose a novel real-time single stage 3D object detector, namely CVFNet, which is able to learn and fuse cross-view features more consistently. In addition to common range view and bird's eye view, a simple point view which only concerns with MLP operations is induced in our network to reduce the information loss. We transform views and fuse features in a progressive manner to relief feature inconsistency~\cite{wang2021pointaugmenting}. 
We first propose a Point-Range fusion module (PRNet) to extract and fuse point-view and range-view features with efficient feature interaction. By this means, we obtain 3D point-wise features with well-preserved 3D geometric and 2D textural information. Next, we design a Slice Pillar transformation module to transform the point-view features into BEV features. To further improve the performance and save the computational cost, a Sparse Pillar Head is utilized to filter out the empty pillars. We conduct experiments on the KITTI~\cite{Geiger2013VisionMR} and NuScenes~\cite{Caesar2020nuScenesAM} dataset and state-of-the-art performances are achieved among the view-based methods. 
Meanwhile, our framework features utilizing only 2D convolutions (including MLPs), leading to high real-time performance close to PointPillars~\cite{Lang_2019_CVPR}.

The main contributions can be summarized as follows:
\begin{itemize}
\item A novel real-time view-based one stage 3D object detector is proposed. It comprises only 2D convolutions and is able to learn consistent cross-view features through a progressive fusion manner. 
\item A Point-Range fusion module PRNet is designed to effectively extract and fuse the 3D point and 2D range view features in a bilateral aggregation way.
\item A Slice Pillar transformation module is presented to transform 3D point features into 2D BEV features, which features less 3D geometric information loss during the dimension collapse.
\item Our proposed CVFNet model outperforms existing view-based detectors on KITTI and NuScenes dataset and also shows competitive performance to point or voxel based methods.
\end{itemize}

\section{Related Work}\label{sec:relatedwork}
This section reviews the 3D object detection approaches according to the base representations of 3D point clouds. The methods can be roughly classified into three categories: point-based, voxel-based and view-based methods.

\textbf{Point-based 3D Object Detector.}
The pioneering work PointNet~\cite{Qi_2017_CVPR} and PointNet++~\cite{Qi2017PointNetDH} which extract point-wise feature directly from point cloud also inspire a lot of 3D object detection methods~\cite{Qi2018FrustumPF, Shi2019PointRCNN3O, Shi2020PointGNNGN, Wang2019FrustumCS, Luo20203DSSDLH}. 
F-PointNet~\cite{Qi2018FrustumPF} employs 2D detectors to narrow the search space of points to a frustum and regresses 3D bounding box within it. 
PointRCNN~\cite{Shi2019PointRCNN3O} leverages PointNet++ \cite{Qi2017PointNetDH} with the set abstraction layer and feature propagation layer to segment foreground points from raw point clouds to reduce the searching space, and predicts 3D bounding boxes for each instance point. 
Point-based methods are relatively slow due to the ball query operation in feature propagation layer. Instead of using the propagation layers, we propose PR-fusion blocks to aggregate local features and expand the receptive field for point-wise features.

\begin{figure*}[h]
   \setlength{\abovecaptionskip}{-0.1cm}
   \centering
   \begin{tabular}{ccc}
   \includegraphics[width=17cm]{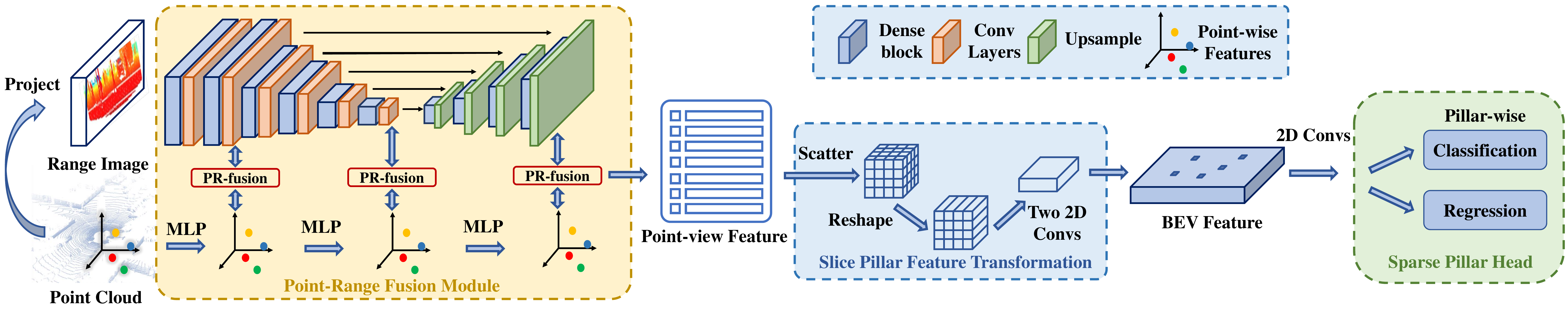}
   \end{tabular}
   \caption{Illustration of the architecture of CVFNet. The features are learned and transformed progressively in 'Range – Point – BEV' order along which the Point view serves as reference.}
   \vspace{-0.4cm} 
   \label{fig:pipeline}
   
\end{figure*}

\textbf{Voxel-based 3D Object Detector.}
Voxel-based methods divide the point cloud into voxels and rely on 3D convolutions for feature learning. In VoxelNet~\cite{Zhou2018VoxelNetEL}, points are equally grouped into voxels and passed to feature encoding layers to predict 3D boxes. SECOND~\cite{Yan2018SECONDSE} designs an effective sparse convolution algorithm by encoding valid voxels only, which improves the running speed significantly.
Voxel R-CNN \cite{Deng2021VoxelRT} takes full advantage of voxel features in a two-stage approach and verifies that coarse voxel granularity can also offer sufficient accuracy. These voxel-based methods are generally effective for accurate 3D object detection, while at the cost of using computation-intensive 3D convolution.

\textbf{View-based 3D Object Detector.}
This category includes bird’s eye view (BEV) based and range view based methods. 
BEV-based methods attract much attention since it features no scale variation for scenes. Pixor~\cite{Yang2018PIXORR3} proposes a novel loss function for 3D object detection on BEV map. 
PointPillars~\cite{Lang_2019_CVPR} directly applies PointNet~\cite{Qi_2017_CVPR} to collapse the height dimension and use it as a pseudo-image for detection. 
Range view is another popular representation with minor information loss, which is consistent with the point production pattern of rotated LiDAR sensors. LaserNet~\cite{Meyer2019LaserNetAE} exploits the range-view in their own dataset and predicts a multimodal distribution for each point. 
There are also methods integrating multiple views to improve the detection performance. MV3D~\cite{Chen2017Multiview3O} fuses the features from BEV, range view as well as RGB images. AVOD \cite{Ku2018Joint3P} expands MV3D by exploiting the 3D anchors into BEV and RGB images. 
These methods mainly extract view features independently and aggregate them at proposal stage, which may cause inconsistency among features.
In contrast, our method learns cross-view features in a progressive way, \textit{i},.\textit{e}., transforming features in sequential views so that the feature inconsistency problem can be largely relieved. 

\section{Method}\label{sec:method}

In this section, we describe our CVFNet model in detail. As illustrated in Fig \ref{fig:pipeline}, the framework mainly comprises the Point-Range fusion module, Slice Pillar transformation, and Sparse Pillar Head. Given the input of point and range view, the features are learned and aggregated effectively through 'Range - Point - BEV' views in a progressive way. Rather than 3D convolutions or time-consuming ball query point operations, our model involves only fast 2D convolutions and MLPs, which guarantee the real-time performance of the system.


\subsection{Point-Range Fusion Module}\label{sec:prnet}
Range view is the native form of rotating LiDAR, featuring dense range images distributed in spherical coordi-nates where 2D convolutions can be applied effectively. Point view is of 3D coordinates that multi-layer perceptron (MLPs) can be used to extract features for each point. To extract and aggregate the features of these two views, we propose a Point-Range fusion module within which multi-stage bilateral fusion is carried out.

\textbf{Point-Range Mapping.} Given a 3D point cloud as input, we first transform it into range representation as in~\cite{Milioto2019RangeNetF}. Each point $\textbf{p}_i=(x, y, z)$ is converted via a mapping $\Pi$ : $ \mathbb{R}_3 \mapsto \mathbb{R}_2 $ to spherical coordinates, which are then scaled as planar image coordinates $(u,v)$ within an image of size $h \times w$. The process is described as:
$$
\dbinom{u}{v}=\binom{\frac{1}{2}[1-\rm{arctan}(\it{y}, \it{x})\pi^{-1}] \it{w}}{[1-(\rm{arcsin}(\it{z}\it{r}^{-1})+\rm{fov}_{up})\rm{fov}^{-1}]\it{h}}
$$

where $\rm{(h,w)}$ are the height and width of the desired range image, $\rm{fov}$ and $\rm{fov_{up}}$ are the entire and upper part of vertical field-of-view of the sensor respectively, and $r =||\textbf{p}_i||_2$ is the range of each point. This procedure produces a point-to-pixel index table, which can help transform the features between point and range view.

\begin{figure}[t]
   \centering
   \begin{tabular}{ccc}
   \includegraphics[width=8cm]{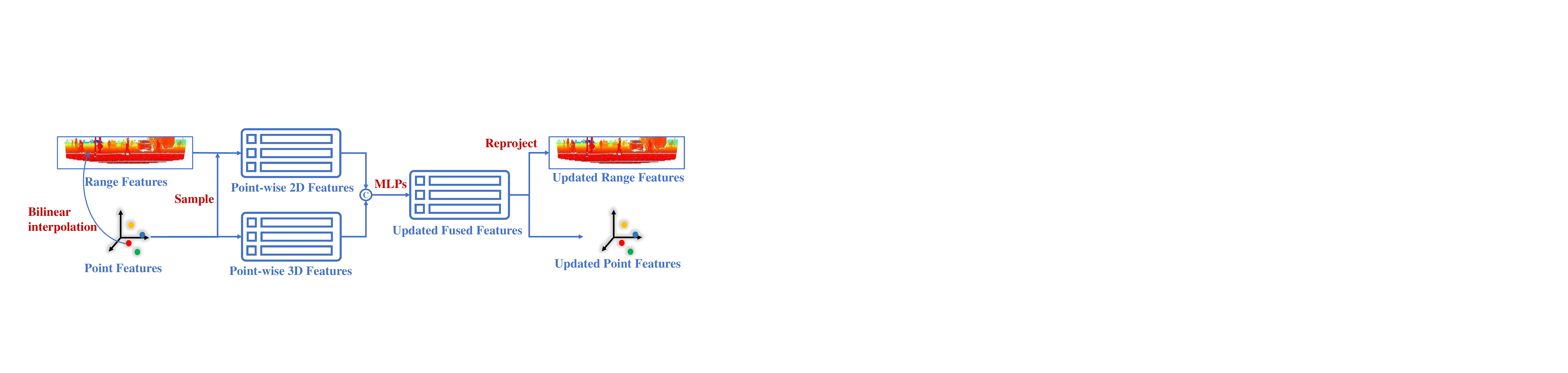}
   \end{tabular}
   \caption{Our PR-fusion block to update both Point and Range features.}
   \label{fig:transformation}
\end{figure}

\textbf{Module Structure.} As shown in Fig~\ref{fig:pipeline}, the Point-Range fusion module comprises two main streams, \textit{i}.\textit{e}., Point stream and Range stream, interactively connected by three PR-fusion blocks. The Range stream is an encoder-decoder architecture based on 2D denseblocks~\cite{Huang2017DenselyCC}. In the encoder stage, we insert denseblocks with a stride of 2 to gradually produce 2x, 4x, 8x, 16x downscaled feature maps. While in the decoder stage, we apply denseblocks with a stride of 1, and use bilinear interpolation to gradually recover the original size. To better utilize the encoder features, we directly concatenate the encoder features into the the decoder stage as in U-Net~\cite{Ronneberger2015UNetCN}. For the Point stream, we just use several PointNet-like MLPs (a series of fully-connected layers) for feature extraction. We do not use any complex set abstraction layers or feature propagation layers in ~\cite{Qi2017PointNetDH} to avoid time-consuming operations. 

\textbf{PR-fusion Block.} The PR-fusion block is responsible for deeply fusing and propagating the features between point and range view. The detailed structure of this block is illustrated in Fig \ref{fig:transformation}. Given the range and point view features in a certain stage of the main stream, we first sample 2D range features for each 3D point by searching the point-range indexing table. To deal with the scale variation, we use bilinear interpolation to get the features at the right scale. And for point-view features, we directly get the point-wise 3D features from the feature map. Then, we concatenate these point-wise 2D and 3D features and pass them to a series of MLPs. The resulting aggregated features are reprojected back to the range and point view respectively to continue the subsequent process of the network. The PR-fusion block is applied three times in the module, corresponding to early, middle and late stage of fusion, respectively.

\begin{figure}
  \setlength{\abovecaptionskip}{-0.15cm}
   \centering
   \begin{tabular}{ccc}
   \includegraphics[width=8cm]{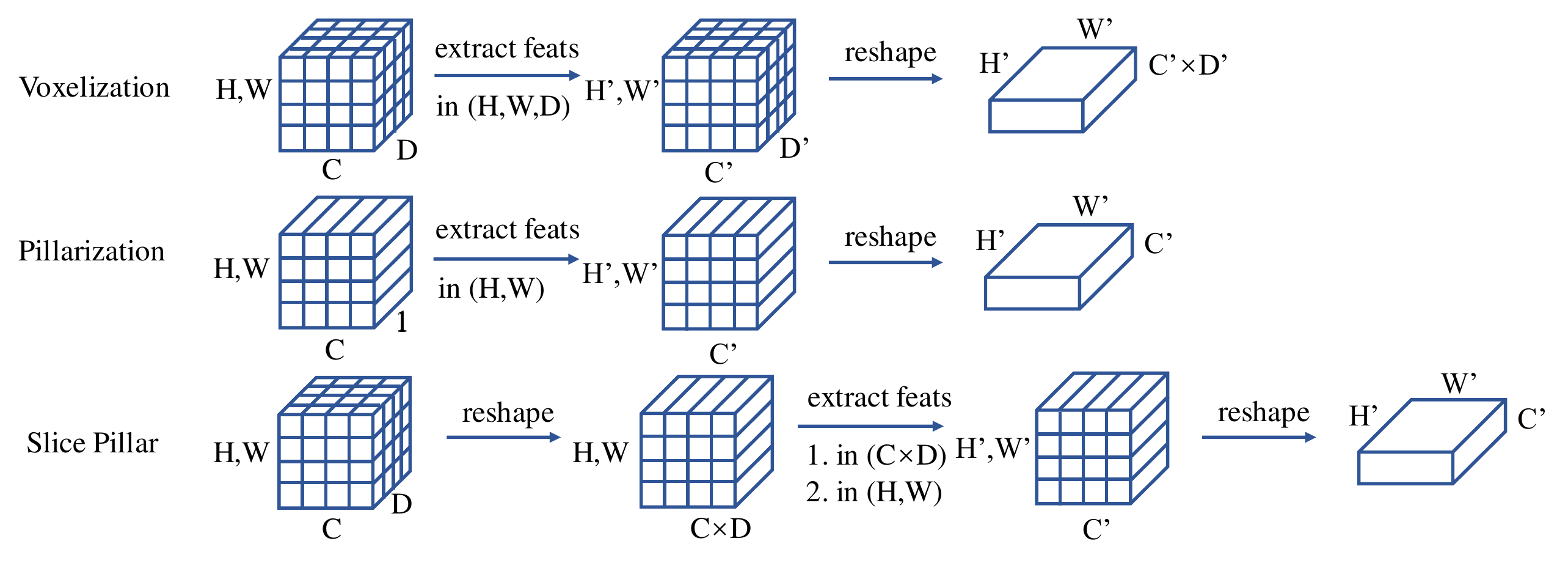}
   \end{tabular}
   \caption{The Comparison of Point-to-BEV Feature Transformation Modules.}
   \label{fig:slicepillar}
\end{figure}

\subsection{Slice Pillar Transformation}\label{sec:slicepillar}
Bird's eye view(BEV) is a popular representation in 3D object detection as it features no occlusion or scale variation. There are two common ways to project the point cloud to BEV: voxelization and pillarization.

Given a voxelized point cloud with the shape of $\rm{(H, W, D, C)}$, voxel-based methods extract features by 3D convolutions in $\rm{H, W}$ and $\rm{D}$ dimensions, where $\rm{D}$ represents height dimension and $\rm{C}$ represents feature channels. Downsampled features with the size of $\rm{(H^{'}, W^{'}, D^{'}, C^{'})}$ are then reshaped into BEV features with size $\rm{(H^{'}, W^{'}, D^{'} \times C^{'})}$. On the other hand, pillar-based methods directly use PointNet to extract features for points in single pillars with size $\rm{(H, W, 1, C)}$, which has no feature resolution on height dimensions. It is then formatted as $\rm{(H^{'}, W^{'}, C^{'})}$ on the BEV perpective before feeding it into the detection head. While voxel-based methods maintain smaller feature granularity, they rely on 3D convolutions which are 
computation intensive. Pillar-based methods \cite{Lang_2019_CVPR} are more computationally efficient, while they suffer information loss along the height dimension.

To tackle these problems, we design an improved version of the pillar strategy, namely Slice Pillar. The differences to the previous two transformations are shown in Fig~\ref{fig:slicepillar}.  We first calculate the 3D index of each point and use scatter operations, which has a much higher speed than voxel generator, to transform the point-wise features into voxel-wise feature volume with size of $\rm{(H, W, D, C)}$. Scatter operation is implemented in Pytorch and will pick one point arbitrarily while propagating the gradient to points that have the same voxel index. Then we combine the last two dimensions of features, resulting in a feature volume of $\rm{(H, W, D \times C)}$. Upon these sliced pillars, a couple of fully-connected layers (or $1\times1$ convolutions) are applied along $\rm{(D \times C)}$ dimension, followed by a series of common 2D convolutions along BEV’s planar $\rm{(H, W)}$ dimensions. By this means, fine granularity features in all three dimensions can be extracted without using 3D convolutions.

\begin{table*}[t]
   \begin{center}
   \caption{Comparisons of 3D object detection on KITTI \textit{test} set. The best scores are marked in bold with '*' indicates RGB images integrated. All runtime values are collected from KITTI benchmark as well as official paper and code.}
   \label{table:kittitest}
   \scalebox{0.9}{
   \begin{tabular}{c|c|c|c|c|c|c|c|c|c}
   \toprule
   \multirow{2}{*}{Method} & \multirow{2}{*}{Representation} &  \multirow{2}{*}{Stage} & \multicolumn{3}{|c|}{3D AP}  &\multicolumn{3}{|c|}{BEV AP} & \multirow{2}{*}{Time(ms)} \\ 
   \cmidrule{4-9}
   &  &  & Easy & Moderate & Hard & Easy & Moderate & Hard  \\
   \midrule
   F-PointNet$\rm{^*}$ \cite{Qi2018FrustumPF}  & Point & Multi & 82.19 & 69.79 & 60.59 & 91.17 & 84.67 & 74.77 & 170\\  
   PI-RCNN$\rm{^*}$ \cite{Xie2020PIRCNNAE} & Point & Two & 84.37 & 74.82 & 70.03 & 91.44 & 85.81 & 81.00 & 100\\
   PointRCNN \cite{Shi2019PointRCNN3O} & Point & Two & 86.96 & 75.64 & 70.70 & 92.13 & 87.39 & 82.72 & 100 \\
   F-ConvNet$\rm{^*}$ \cite{Wang2019FrustumCS} & Point & Multi & 87.36 & 76.39 & 66.69 & 91.51 & 85.84 & 76.11 & 470 \\
   PointGNN \cite{Shi2020PointGNNGN} & Point & Two & 88.33 & 79.47 & 72.29 & \textbf{93.11} & \textbf{89.17} & 83.90 & 643 \\
   3D-SSD \cite{Luo20203DSSDLH} & Point & One & \textbf{88.36} & \textbf{79.57} & \textbf{74.55} & 92.66 & 89.02 & \textbf{85.86} & \textbf{38} \\
   \midrule
   VoxelNet \cite{Zhou2018VoxelNetEL} & Voxel & One & 77.82 & 64.17 & 57.51 & 87.95 & 78.39 & 71.29 & 220 \\
   UberATG-ContFuse$\rm{^*}$ \cite{Liang2018DeepCF} & Voxel & One & 83.68 & 68.78 & 71.67 & 94.07 & 85.35 & 75.88 & 60 \\
   SECOND \cite{Yan2018SECONDSE} & Voxel & One & 83.34 & 72.55 & 65.82 & 89.39 & 83.77 & 78.59 & 50 \\
   TANet \cite{Liu2020TANetR3} & Voxel & One & 84.39 & 75.94 & 68.82 & 91.58 & 86.54 & 81.19 & \textbf{34.75} \\
   3D IoU Loss \cite{Zhou2019IoULF} & Voxel & Two & 86.16 & 76.50 & 71.39 & 91.36 & 86.22 & 81.20 & 80 \\
   Fast PointRCNN \cite{Chen2019FastPR} & Voxel & Two & 85.29 & 77.40 & 70.24 & 90.87 & 87.84 & 80.52 & 65 \\
   Associate-3Ddet \cite{Du2020Associate3DdetPA} & Voxel & One & 85.99 & 77.40 & 70.53 & 91.40 & 88.09 & 82.96 & 60 \\
   PV-RCNN \cite{Shi2020PVRCNNPF} & Voxel & Two & 90.25 & 81.43 & 76.82 & \textbf{94.98} & \textbf{90.65} & \textbf{86.14} & 80  \\
   Voxel-RCNN \cite{Deng2021VoxelRT} & Voxel & Two & \textbf{90.90} & \textbf{81.62} & \textbf{77.06} & 94.85 & 88.83 & 86.13 & 40  \\
   \midrule
   MV3D$\rm{^*}$ \cite{Chen2017Multiview3O} & View & Two & 74.97 & 63.63 & 54.00 & 86.62 & 78.93 & 69.80 & 360  \\
   AVOD-FPN$\rm{^*}$ \cite{Ku2018Joint3P} & View & Two & 83.07 & 71.76 & 65.73 & 90.99 & 84.82 & 79.62 & 100 \\
   UberATG-Pixor++ \cite{Yang2018PIXORR3} & View & One & - & - & - & 93.28 & 86.01 & 80.11 & 35  \\
   MoDet \cite{Zhang2019AccurateAR} & View & One & - & - & - & 90.80 & 87.56 & 82.69 & 50 \\
   PointPillars \cite{Lang_2019_CVPR}  & View & One & 82.58 & 74.31 & 68.99 & 90.07 & 86.56 & \textbf{82.81} &  \textbf{23.6} \\
   LaserNet \cite{Meyer2019LaserNetAE}  & View & One & - & - & - & 79.19 & 74.52 & 68.45 & 30 \\
   MAFF-Net$\rm{^*}$ \cite{Zhang2020MAFFNetFF}  & View & One & 85.52 & 75.54 & 67.61 & 90.79 & 87.34 & 77.66 & 40 \\
   SARPNET \cite{Ye2020SARPNETSA}  & View & One & 85.63 & 76.64 & 71.31 & 92.21 & 86.92 & 81.68 & 50 \\
   \midrule
   Ours  & View & One & \textbf{88.75} & \textbf{77.70} & \textbf{71.95} & \textbf{93.65} & \textbf{87.87} & 82.29 &  \textbf{28.1} \\
   \bottomrule
   \end{tabular}}
  \vspace{-0.3cm} 
   \end{center}
   
\end{table*}


\begin{table}[t]
    \setlength{\abovecaptionskip}{-0.4cm}
   \caption{Performance Comparision of $\rm{AP_{3D}} $ in other categories on the KITTI \textit{validation} set.}
   \label{table:kittival}
   \begin{center}
   \scalebox{1.0}{
   \begin{tabular}{c|c|c|c|c|c|c}
  \toprule
    \multirow{2}{*}{Method} & \multicolumn{3}{|c|}{Pedestrian}  &\multicolumn{3}{|c}{Cyclist}  \\ 
   \cmidrule{2-7}
    & Eas. & Mod. & Har. & Eas. & Mod. & Har.  \\
   \midrule
   PointPillars~\cite{Lang_2019_CVPR} & 54.18 & 49.70 & 45.00 & 80.38 & 63.29 & 60.05  \\
   PFFNet~\cite{Fu2021ImprovedPW} & 55.46 & 50.09 & 45.49 & - & - & - \\
   \midrule
   Ours & \textbf{60.79} & \textbf{53.95} & \textbf{49.04} & \textbf{83.07} & \textbf{65.07} & \textbf{61.34}  \\
   \bottomrule
   \end{tabular}}
  \vspace{-0.6cm} 
   \end{center}
\end{table}

\subsection{Sparse Pillar Head}\label{sec:pillarhead}
After the slice pillar transformation module and a series of convolution layers on BEV, we have the BEV feature map at hand. To estimate the proposals in each BEV grid, most existing methods use a dense prediction head. However, since most grids are empty, we introduce a sparse pillar head to suppress the useless proposals. The sparse pillar head consists of classification and regression head. Given a sparse BEV map, we first define nonempty grids as valid pillars and use 2D convolutions to extract features, which are then indexed to the pillar-wise features. Finally, classification head is used to predict the probability of a pillar’s class, and regression head is responsible for regressing 3D box parameters. The advantages of our sparse pillar head are two folds. Firstly, it balances the distribution of positive and negative samples by filtering out the empty grids which are otherwise counted as negative samples. Secondly, the computation cost is further reduced. 

\subsection{Loss Function}\label{sec:loss}
The overall loss for KITTI dataset consists of pillar-wise anchor classification loss $\rm{L_{cls}}$, regression loss $\rm{L_{reg}}$ and direction classification loss $\rm{L_{dir}}$. The final loss is the weighted sum of three losses:
$$
\rm{Loss} = \rm{L_{cls}} + \alpha \rm{L_{reg}} + \beta \rm{L_{dir}}
$$
where $\alpha$ is set to 2.0 and $\beta$ is set to 0.2. We use the focal loss with default parameters as \cite{Lin2017FocalLF} for $\rm{L_{cls}}$ and $\rm{L_{dir}}$. For $\rm{L_{reg}}$, the smooth-L1 loss is employed to calculate the residual value relative to the predefined anchor of ground truth boxes. 

In NuScenes dataset, we use center-based anchor-free head to be adaptive to the requirement of 10 categories detection. The loss follows CenterPoint~\cite{Yin2021Centerbased3O} as:
$$
\rm{Loss} = \rm{L_{hm}} + \lambda_{box} \rm{L_{box}} + \lambda_{local} \rm{L_{local}}
$$
where $\rm{L_{hm}}$ is the loss of heatmap for center classification, $\rm{L_{box}}$ is the loss for regressing box parameters, $\rm{L_{local}}$ is the loss for regressing local offset and $\lambda$s are the related loss weights, which are both set to 0.25. 

\section{Experiments}\label{sec:experiments}
We conduct experiments on the KITTI and NuScenes dataset. We describe the dataset settings in Sec \ref{sec:dataset} and implementation details in Sec~\ref{sec:implementation}. In Sec~\ref{sec:kittitest}, we compare our CVFNet with the state-of-the-art 3D object detection methods. Ablation studies are conducted in Sec \ref{sec:ablation}.

\subsection{Dataset and Evaluation}\label{sec:dataset}
We compare the quantitative performance of our method on well-known KITTI and NuScenes dataset. 

The KITTI~\cite{Geiger2013VisionMR} 3D object detection dataset contains 7481 training samples and 7518 test samples. The training set is divided into 3712 samples for training and 3769 samples for validation\cite{Chen20153DOP}. The samples are classified as Car, Pedestrian and Cyclist with three difficulty levels for each class: Easy, Moderate, and Hard. The official KITTI leaderboard is ranked on Moderate difficulty. CVFNet is evaluated on the KITTI \textit{test} set by submitting the detection results to the official server. We use average precision (AP) with an IoU threshold 0.7 for Car and 0.5 for both Pedestrian and Cyclist. 

The NuScenes~\cite{Caesar2020nuScenesAM} 3D detection dataset contains 1000 scenes, including 700 scenes for training, 150 scenes for validation and 150 scenes for testing. Each scene is of 20{\it s} duration and captured by a 32-beam LiDAR. There are up to 1.4M annotated 3D bounding boxes for 10 classes: car, truck, bus, trailer, construction vehicle, pedestrian, bicycle, barrier and traffic cone. The mAP is for measuring the precision and recall which is defined based on the match by 2D center distance on the ground plane instead of IoU. NDS (Nuscenes Detection Score) is the composite score in terms of the precision of translation, scale, orientation, velocity and attributes.  

\subsection{Implementation Details}\label{sec:implementation}

For Point-Range fusion module, we project the raw point cloud into range image with the size of $48\times512\times5$ as \cite{Liang2020RangeRCNNTF} in KITTI and $64\times2048\times6$ in NuScenes dataset, respectively. The input five channels for each pixel include range, 3D x, y, z coordinates and intensity. An extra timestamp channel is added in NuScenes dataset as multi-sweep strategy is applied. 
The backbone for range stream is an encoder-decoder structure with 10 denseblocks. The backbone network extracts range features with strides of (1,1,2,2,2,2) and 4 denseblocks with bilinear interpolation operations to gradually recover the original size. We insert 6 convolution layers after each denseblock in encoders with the number of (3,3,5,5,5,5) layers to extract explicit features. For the MLPs in Point stream, we apply a couple of fully connected layers like PointNet \cite{Qi_2017_CVPR} to extract 3D point-wise features.  For the detection head, we use an anchor-based head \cite{Lang_2019_CVPR} in KITTI dataset and an anchor-free head \cite{yin2021center} in NuScenes dataset.

For the scatter operations in KITTI dataset, we define the detection range as [0, 69.12]m for X axis, [-39.68, 39.68]m for Y axis and [-3, 1]m for Z axis. We set the scatter voxel size to (0.16, 0.16, 0.2) and the final 3D volumes is of size with  $\rm{H=496, W=432, D=20}$ and $\rm{C=64}$, respectively. In Nuscenes dataset, the detection range is $[-51.2, 51.2]$m for the X,Y axis and [-3, 5]m for the Z axis. The voxel size is set to (0.2, 0.2, 0.4) and the final 3D volumes is of size with  $\rm{H=512, W=512, D=20}$ and $\rm{C=64}$.
After Point-to-BEV transformation, we use three 2D convolution blocks with stride of 2 to downscale features and then upscale each output to $\frac{1}{2}$ of the original size, The maps are finally concatenated together to get the BEV features for the detection head.

\begin{table*}[t]
\begin{center}
\caption{Comparisons of 3D object detection on NuScenes \textit{test} set.}
\label{table:nus_test}
\begin{tabular}{c|c|c|c|c|c|c|c|c|c|c|c|c}
\toprule
Method & mAP & NDS & Car & Truck & Bus & Trailer & Cons. & Pedes. & MCycle & bicycle & cone & barrier \\ 
\midrule
PointPillars~\cite{Lang_2019_CVPR} & 30.5 & 45.3 & 68.4 & 23.0 & 28.2 & 23.4 & 4.1 & 59.7 & 27.4 & 1.1 & 30.8 & 38.9 \\
SARPNET~\cite{Ye2020SARPNETSA} & 31.6 & 49.7 & 59.9 & 18.7 & 19.4 & 18.0 & 11.6 & 69.4 & 29.8 &  14.2 & 44.6 & 38.3 \\
CBGS~\cite{Zhu2019ClassbalancedGA} & 52.8 & \textbf{63.3} & 81.1 & \textbf{48.5} & 54.9 & 42.9 & 10.5 & 80.1 & 51.5 & 22.3 & 70.9 & 65.7 \\
3D-CVF~\cite{Yoo20203DCVFGJ} & 52.7 & 62.3 & 83.0 & 45.0 & 48.8 & 49.6 & \textbf{15.9} & 74.2 & 51.2 & 30.4 & 62.9 & 65.9 \\
PanoNet3D~\cite{Chen2020PanoNet3DCS} & 54.5 & 63.1 & 80.1 & 45.4 & 54.0 & 51.7 & 15.1 & 79.1 & \textbf{53.1} & \textbf{31.1} & 71.9 & 62.9 \\
\midrule
CVFNet & \textbf{54.9} & \textbf{63.3} & \textbf{83.0} & 47.9 & \textbf{56.3} & \textbf{53.0} & 13.7 & \textbf{80.4} & 46.7 & 27.5 & \textbf{72.4} & \textbf{67.8} \\

\bottomrule
\end{tabular}
\vspace{-0.3cm} 
\end{center}
\end{table*}

\begin{table*}[t]
\begin{center}
\caption{Effects of Proposed Components for 'Car' Detection on KITTI \textit{validation} set.}
\label{table:components}
\begin{tabular}{c|c|c|c|c|c|c|c|c|c}
\toprule
\multirow{2}{*}{Baseline} & \multirow{2}{*}{PRNet} &  \multirow{2}{*}{Slice Pillar} & \multirow{2}{*}{Sparse Head} & \multicolumn{3}{|c|}{3D}  &\multicolumn{3}{|c}{BEV} \\ 
\cmidrule{5-10}
&  &  &  & Easy & Moderate & Hard & Easy & Moderate & Hard  \\
\midrule
\checkmark & & & &             87.66 & 76.01 & 73.38 & 92.62 & 87.38 & 83.89 \\
\checkmark & \checkmark & & &  88.99 & 78.64 & 76.19 & 93.16 & 88.42 & 85.43 \\
\checkmark & & \checkmark & &  89.72 & 79.37 & 75.44 & 93.67 & \textbf{88.98} & 86.31 \\
\checkmark & & & \checkmark &  87.62 & 77.13 & 74.77 & 92.91 & 88.30 & 83.98 \\
\checkmark & \checkmark & \checkmark & &  89.91 & 79.49 & 76.42 & 93.84 & 88.90 & \textbf{86.62} \\
\checkmark & \checkmark & \checkmark & \checkmark & \textbf{90.02} & \textbf{79.82} & \textbf{76.84} & \textbf{93.89} & 88.96 & 86.50 \\

\bottomrule
\end{tabular}
\vspace{-0.6cm} 
\end{center}
\end{table*}

\begin{table}[t]
\setlength{\belowcaptionskip}{+0.5cm}
\begin{center}
\caption{Effects of Proposed Components on NuScenes \textit{validation} set.}
\label{table:nus_components}
\begin{tabular}{c|c|c|c|c|c}
\toprule
Baseline & PRNet & Slice Pillar & Sparse Head & mAP & NDS \\
\midrule
\checkmark & & & & 49.60 & 60.33 \\
\checkmark & \checkmark & & & 51.89 & 61.48 \\
\checkmark & & \checkmark & & 51.13 & 60.95\\
\checkmark & & & \checkmark & 49.99 & 60.19 \\
\checkmark & \checkmark & \checkmark & & 53.21 & 61.97 \\
\checkmark & \checkmark & \checkmark & \checkmark & \textbf{53.46} & \textbf{62.09} \\

\bottomrule
\end{tabular}
\vspace{-0.3cm} 
\end{center}
\end{table}

We train our network end-to-end with the ADAM optimizer of a batch size 12/8 on two NVIDIA RTX2080Ti GPUs for 80/20 epochs on KITTI/NuScenes dataset, respectively. The learning scheme follows cyclic policy, and the learning rate changes from 0.001 to 0.01, then goes to $0.001\times10^{-4}$. The momentum ranges from 0.85 to 0.95. 

For data augmentation, we use the random flipping along the \textit{x} axis, random global scaling with the scaling factor in $[0.95,1.05]$ and random global rotation ranges from $[-\frac{\pi}{4}, \frac{\pi}{4}]$ around the \textit{z} axis. Some objects are also randomly sampled from the training data and injected into training samples.

\subsection{Comparison with State-of-the-Arts}\label{sec:kittitest}
\textbf{KITTI} We evaluate our CVFNet detector on the KITTI \textit{test} set and submit the detecting results to the official online benchmark server. The mAP performance calculated with 40 recall positions on car category is recorded in Table~\ref{table:kittitest}.
In the context of view-based detectors, our method outperforms all of its competitors by a noticeable margin. In particular, the 3D AP of our model outperforms the baseline method vanilla PointPillars~\cite{Lang_2019_CVPR} by 6.17\%, 3.39\% and 1.96\% in three difficulty levels respectively. 
With similar pillar representations on the BEV, this large improvement mainly comes from our specially designed cross view feature learning  scheme. Besides, our model even outperforms most of the point-based and voxel-based methods, which verifies the benefits of exploiting 2D view features from 3D point clouds.

\begin{figure*}[ht]
\setlength{\abovecaptionskip}{-0.1cm}
\centering
\begin{tabular}{ccc}
\includegraphics[width=16cm]{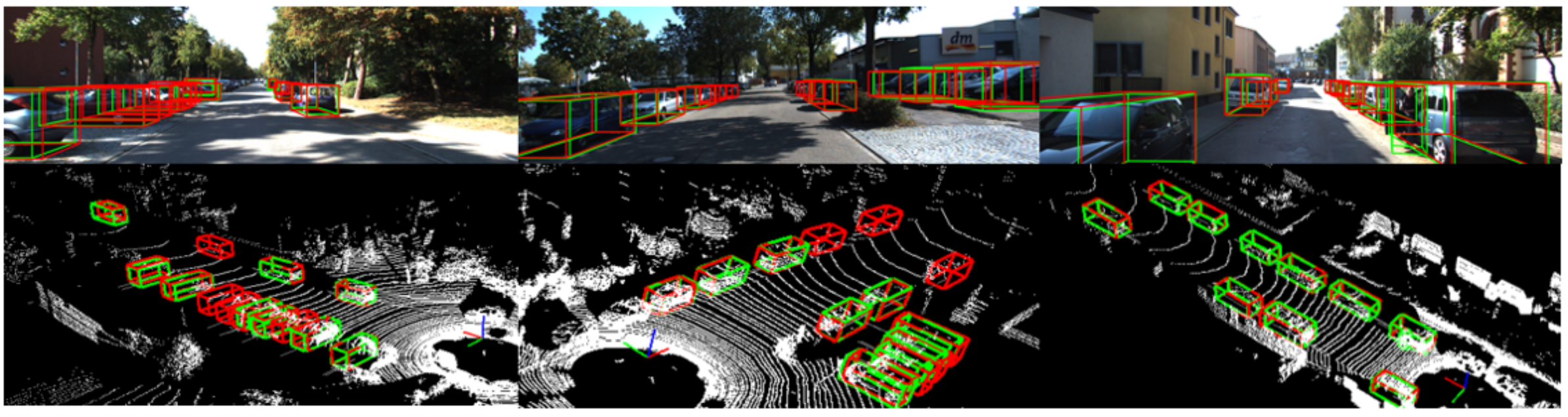}
\end{tabular}
\caption{The detection results on KITTI \textit{validation} set. We render the predicted and ground truth bounding boxes in red and green, respectively.}
\label{fig:detection} 
\vspace{-0.4cm}
\end{figure*}

We also show the results in three class on KITTI \textit{validation} set in Table~\ref{table:kittival}. Results are evaluated by the mAP with 11 recall positions. Compared with the baseline method PointPillars \cite{Lang_2019_CVPR}, our method achieves about 4.25\% and 1.78\% improvement on moderate AP for Pedestrian and Cyclist respectively, which further depicts the superior generalization of our method. 
Some qualitative results are illustrated in Fig~\ref{fig:detection}, where the detected 3D boxes are also projected to the RGB images for better visualization.

\textbf{NuScenes} 
To test the robustness of our CVFNet, we also evaluate our method on the NuScenes dataset and submit the detection results to the official server. The mAP and NDS performance of 10 categories is listed in Table \ref{table:nus_test}. Compared with other view-based methods, our method get the highest AP in most categories, which indicates our method's robustness for different scenes and objects.

\textbf{Timing}
Owing to the fast 2D CNN backbone and one-stage architecture, our CVFNet achieves 35.6 Hz in KITTI and 16.4 Hz in NuScenes when evaluated on a single 2080Ti GPU without any C++/CUDA/TensorRT optimizations. The comparisons of the running speed in KITTI are shown on the rightest column of Table~\ref{table:kittitest}, where our algorithm ranks the second fastest.

\subsection{Ablation Studies}\label{sec:ablation}
We conduct ablation studies on both KITTI and NuScenes datasets. 

\textbf{Effects of Components}
We study the effects of the different components of CVFNet by setting the the vanilla PointPillars \cite{Lang_2019_CVPR} as the baseline. The results on KITTI dataset are shown in Table \ref{table:components}. With the PRNet on (Row 2), the 3D and BEV AP has a 1\%-3\% improvement. Adding the Slice Pillar and Sparse Head alone also have positive effects on the baseline. Adding all of the three components can achieve the best results. The ablation results on Nuscenes shown in Table~\ref{table:nus_components} also show the similar effects. 

\begin{table}[t]
   \begin{center}
   \caption{Effects of different view features}
   \label{table:features}
   \scalebox{0.9}{
   \begin{tabular}{c|c|c|c|c|c|c|c}
   \toprule
   \multirow{2}{*}{Point} & \multirow{2}{*}{Range} & \multicolumn{3}{|c|}{3D}  &\multicolumn{3}{|c}{BEV} \\ 
   \cmidrule{3-8}
      &  & Easy & Moderate & Hard & Easy & Moderate & Hard  \\
   \midrule
   \checkmark &           & 87.68 & 77.55 & 73.34 & 92.59 & 88.54 & 84.07 \\
   & \checkmark           & 89.11 & 78.98 & 75.27 & 92.62 & 88.46 & 85.49 \\
   \checkmark &\checkmark & 90.02 & 79.82 & 76.84 & 93.89 & 88.96 & 86.50 \\
   \bottomrule
   \end{tabular}}
  \vspace{-0.2cm} 
   \end{center}
\end{table}

\textbf{Effects of different view features}
We analyze the effectiveness of fusing point and range views within the Point-Range fusion module. The experimental results on KITTI dataset are shown in Table \ref{table:features}. We can see that the range view alone performs better than the point view. The combination of both views improves the 3D AP by more than 2\%, which indicates the success of our point and range view fusion strategy.



\section{Conclusion}\label{sec:conclusion}
In this paper, we present a cross-view learning based single-stage 3D object detection method which features both high detection and real-time performance. To better preserve 3D geometric information while maintaining computing efficiency, we design a progressive cross view fusion pipeline to learn and transform the features among Point, Range and BEV views. The special designs of the point-range fusion and the slice pillar module strengthen the solid feature learning. Experimental results achieve state-of-the-art performance among the view-based methods with a high frame rate of 35.6Hz on the KITTI dataset. The simple backbone and quick runtime of our CVFNet make it easy to deploy in resource constrained devices. In future work, we plan to integrate more views sun as RGB image to further improve the performance.


\bibliographystyle{IEEEtran}
\bibliography{IEEEabrv,ref}




\end{document}